\pdfoutput=1
\documentclass[11pt]{article}
\usepackage[]{coling}
\usepackage{mdframed}
\usepackage{listings}
\usepackage{xcolor}
\usepackage[inline]{enumitem}
\usepackage{booktabs}
\usepackage{multirow}
\usepackage{multicol}
\usepackage{tikz}
\usetikzlibrary{positioning} 
\usepackage{amsmath}
\definecolor{verylightgray}{rgb}{0.95,0.95,0.95}
\definecolor{codegray}{rgb}{0.5,0.5,0.5}
\usepackage{times}
\usepackage{latexsym}
\usepackage{booktabs} 
\usepackage[T1]{fontenc}
\usepackage[utf8]{inputenc}
\usepackage{microtype}
\usepackage{inconsolata}
\usepackage{graphicx}

\author{Seyed Amin Tabatabaei\textsuperscript{1}, Sarah Fancher\textsuperscript{2}, Michael Parsons\textsuperscript{2}, Arian Askari\textsuperscript{3} \\
\textsuperscript{1}Elsevier s.tabatabaei@elsevier.com\\
\textsuperscript{2}SSRN \{sarah.fancher, michael\}@ssrn.com\\
\textsuperscript{3}Leiden University a.askari@liacs.leidenuniv.nl \\
}
\newcommand{\header}[1]{\vspace*{1mm}\noindent\textbf{#1}.}
\title{Can Large Language Models Serve as Effective Classifiers for Hierarchical Multi-Label Classification of Scientific Documents at Industrial Scale?}

\begin{document}
\maketitle
\begin{abstract}
We address the task of hierarchical multi-label classification (HMC) of scientific documents at an industrial scale, where hundreds of thousands of documents must be classified across thousands of dynamic labels. The rapid growth of scientific publications necessitates scalable and efficient methods for classification, further complicated by the evolving nature of taxonomies—where new categories are introduced, existing ones are merged, and outdated ones are deprecated. Traditional machine learning approaches, which require costly retraining with each taxonomy update, become impractical due to the high overhead of labelled data collection and model adaptation.
Large Language Models (LLMs) have demonstrated great potential in complex tasks such as multi-label classification. However, applying them to large and dynamic taxonomies presents unique challenges as the vast number of labels can exceed LLMs’ input limits.
In this paper, we present novel methods that combine the strengths of LLMs with dense retrieval techniques to overcome these challenges. Our approach avoids retraining by leveraging zero-shot HMC for real-time label assignment. We evaluate the effectiveness of our methods on SSRN, a large repository of preprints spanning multiple disciplines, and demonstrate significant improvements in both classification accuracy and cost-efficiency.
By developing a tailored evaluation framework for dynamic taxonomies and publicly releasing our code, this research provides critical insights into applying LLMs for document classification, where the number of classes corresponds to the number of nodes in a large taxonomy, at an industrial scale.
\end{abstract}
\footnote{This paper has been accepted at COLING 2025 (Industry Track)}
\section{Introduction}
\label{Introduction} 
The rapid increase in scientific publications presents growing challenges for categorizing these documents in digital repositories. While the volume of papers is significant, the complexity is further increased by the wide range of topics, which are hierarchically organized in a taxonomy since the topics can be viewed as subcategories of broader categories within this hierarchy \cite{liu2023recent,toney2022multi}.
\par
However, taxonomies are not static. Domain experts and librarians frequently update them to reflect advancements in various fields. Categories are regularly introduced, merged, or deprecated to ensure the taxonomy remains up-to-date and relevant. Although HMC has been explored in prior studies, these methods typically assume a fixed taxonomy. To the best of our knowledge, no existing work considers the dynamic nature of taxonomies. 
\par
Given a scientific document and a hierarchical taxonomy of labels, our task is to perform multi-label classification by identifying which leaf node labels from the taxonomy are most appropriate for the document. Current classification approaches, relying on static labels, require retraining whenever the taxonomy changes. This process demands significant amounts of new labeled data given each frequent update of the taxonomy, leading to impractical solutions due to the high cost and time required. Moreover, the large scale of these taxonomies often surpasses the input limitations of most LLMs \cite{chang2024survey, xiong2020approximate,karpukhin2020dense}, which would otherwise be suitable for such complex tasks. 
\par

Label assignment is inherently subjective, as experts may assign different labels to the same document (as illustrated in Figure \ref{fig:perfect_sets} in the Appendix). Our analysis showed that human classification accuracy\footnote{We define human classification as the process of annotating scientific documents under time constraints, which can increase the likelihood of errors due to limited review time. To assess the performance of human classification, a senior and highly experienced subject matter expert annotated the documents with high precision, providing a reference for human accuracy in this context.} varies between 65\% and 90\%, depending on document complexity and taxonomy changes. This inconsistency emphasizes the need for an automated, scalable solution that ensures more consistent and reliable classification results.

\par
In this paper, we propose a novel approach that combines the strengths of LLMs with dense retrieval models. Our methods avoid the high retraining costs associated with machine learning-based approaches by employing zero-shot method for label assignment in large, dynamic taxonomies. We demonstrate the effectiveness of our approach on SSRN, a vast digital repository, showing significant improvements in both accuracy and cost-effectiveness. By automating document categorization, we reduce the costs from \$3.50 per document to approximately 20 cents, marking a pivotal shift for businesses aiming to scale classification efforts while maintaining accuracy.
\par
Our contributions are as follows:
\begin{itemize}
    \item We propose methods for multi-label classification that do not require retraining. These methods leverage LLMs and dense retrieval models to handle large, dynamic taxonomies, making them highly applicable to real-world scenarios where taxonomy structures are periodically evolving.
    \item We introduce a new dataset of scientific documents labeled across multiple disciplines by domain experts. The dataset includes hierarchical, dynamic labels, reflecting the complex structure of modern taxonomies.
    \item We propose a novel evaluation framework tailored to dynamic taxonomies, moving beyond static taxonomies to demonstrate the effectiveness of our methods in a realistic, evolving environment.
    \item We release the code for our methods, enabling reproducibility and fostering future work in HMC with dynamic taxonomies. \footnote{The code and dataset are available at  \url{https://github.com/tabatabaeis/SSRN-LLM-TaxoClass}}
\end{itemize}
\section{Related Work}
HMC of scientific documents has been extensively studied, often with small datasets or static taxonomies \cite{zangari2024hierarchical,wang2024multi,zhu2024hierarchical,zhang2023weakly,fard2023learning,pal2020multi}. 
\par
Previous datasets \cite{kowsari2017HDLTex,lu2003link,yang-etal-2018-sgm,santos2009multi} such as the Cora \cite{cora} and Citeseer \cite{citeseer98} lack hierarchical structures or are limited to a small set of papers. While newer datasets like SciHTC \cite{sadat2022} introduce more hierarchical complexity, they still assume a static taxonomy. 
\par
In our extensive experiments, we found SPECTER2 \cite{singh2022scirepeval} as the most effective baseline on our dataset, which is why we compare our proposed method with it throughout this paper, referring to SPECTER2 as the SOTA on our business-specific dataset. SPECTER \cite{cohan2020specter} processes paper titles and abstracts, optimizing a triplet margin loss that ensures papers with citation links have more similar embeddings than those without. SPECTER2 builds upon this by fine-tuning on four additional tasks: classification, regression, proximity, and retrieval. We further adapt SPECTER2 to our hierarchical multi-label classification task, fine-tuning it for each update of our evolving taxonomy. This process includes manually annotating hundreds of thousands of documents with the new taxonomy labels after each change.
To the best of our knowledge, no prior work has explored the use of LLMs for HMC with either static or dynamic taxonomies. Our work addresses this gap by combining LLMs with dense retrieval models, offering a scalable solution without the need for training.
\section{Dataset Description}
\label{Dataset Description}
\subsection{Document Data}
\label{Document Data}
Document Data includes preprints characterized by title, abstract, and keywords, forming the basis for taxonomy label assignment. See Table \ref{tab:dataset_stats} for the statistics.
In this work, we refer to the preprint or document's `content' as its title, abstract, and keywords.
These features encapsulate the core content and context of each document, serving as the basis for assigning labels from the established taxonomy.
\par
To maintain objectivity and avoid bias, the labelling process excludes author affiliations. For example, a document authored by an individual from a university's law department would not automatically receive labels pertinent to legal studies. This approach ensures that labels are derived solely from the document's content.
While full text is available, it was excluded due to LLM token limits and computational costs, as well as feedback from Subject Matter Experts (SMEs) indicating that manual classification typically relies on metadata alone.
\subsection{Taxonomy Structure}
\label{Taxonomy Structure}
The taxonomy structure is a hierarchical tree with nodes representing scientific disciplines, some with up to nine levels.
The taxonomy is extensive, encompassing several thousand nodes, with some branches extending up to nine levels deep. 
Each node in the taxonomy is defined by its label (name), ID, and its relationships with parent and child nodes.
Additionally, some nodes include a brief description that describes the type of research applicable to that specific node.
The taxonomy is not static; it is regularly reviewed and updated by experts from the repository to reflect the ongoing developments in scientific research. 
This process may involve adding, removing, or merging nodes to ensure the taxonomy remains up-to-date. 
The latest version is available on SSRN\footnote{\url{https://papers.ssrn.com/sol3/DisplayJournalBrowse.cfm}}.
\par
\subsection{Taxonomy Preparation and Enhancement}
\label{taxonomy_prep}
\begin{table}[htbp]
\small
\centering
\caption{Dataset Statistics. }
\begin{tabular}{l|ccc}
\toprule
Statistic & Max & Avg & Min \\ 
\midrule
\multicolumn{4}{c}{Word-level length statistics} \\ 
\midrule
Title & 28 & 13 & 3 \\ 
Keywords & 41 & 9 & 0 \\ 
Abstract & 400 & 180 & 20 \\ 
\midrule
\multicolumn{4}{c}{Hierarchy statistics} \\ 
\midrule
Leaf labels & \multicolumn{3}{c}{\textit{2778 (Total)}} \\ 
Parent labels & \multicolumn{3}{c}{\textit{477 (Total)}} \\ 
Children per parent & 159 & 6.86 & - \\ 
Leaf node depth & 9 & 4.39 & - \\ 
\bottomrule
\end{tabular}
\label{tab:dataset_stats}
\end{table}
\header{Acronym expansion} Our analysis of the taxonomy revealed that many labels contained acronyms, often derived from the names of parent nodes, though some were unrelated. While SMEs generally understand these acronyms, we found that expanding them into full forms enhances LLM comprehension. For instance, FoodSciRN in labels such as \textit{"FoodSciRN Conferences \& Meetings"} refers to \textit{"Food Science Research Network,"} a parent label. Similarly, OPER in labels like \textit{"OPER Subject Matter eJournals"} stands for \textit{"Operations Research Network,"}.
\par
\header{Label description generation}
Our experiments showed that adding label descriptions significantly improved classification effectiveness of various classification approaches. However, manually creating descriptions for around a thousand taxonomy nodes was impractical. To address this, we used GPT-4 to automatically generate descriptions. 
To produce meaningful descriptions, the following information was included in the prompt provided to ChatGPT-4:
(i) Label Name: The name of the node; (ii) Parent Name: The name of the parent node; and (iii) Parent Description: The description of the parent node, if available. The prompt is presented in Figure \ref{fig:descgen}.
We also included a sample description from a node at a similar depth in the taxonomy to guide GPT-4 through few-shot learning. SMEs evaluated the generated descriptions, confirming that most were high quality and suitable for our task. Automating this process enriched the dataset and enhanced the performance of our multi-label classification methods.
\begin{figure*}
    \centering
    \scalebox{0.80}{\includegraphics[width=1\linewidth]{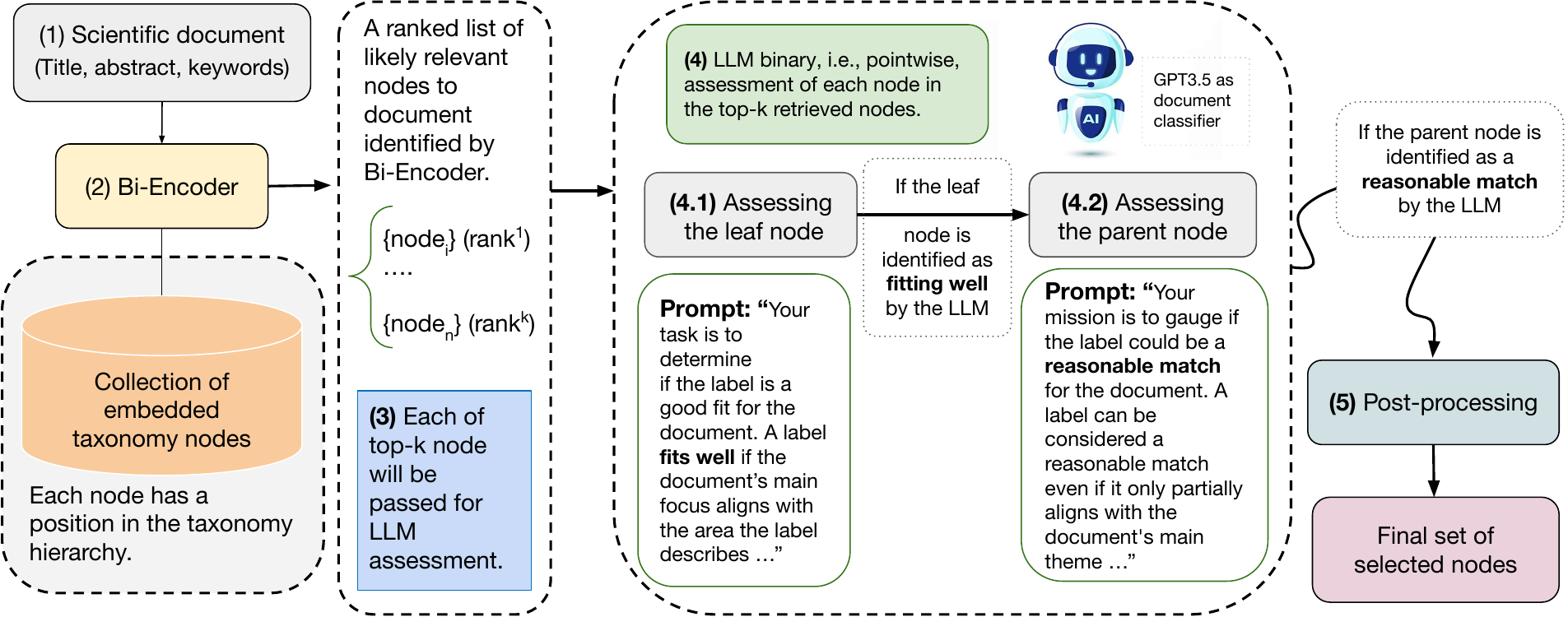}}
    \caption{An illustration of our most effective proposed method, LLM-Select-Pointwise (LLM-SelectP).}
    \label{fig:selectp}
\end{figure*}
\section{Methods}
\label{Methods}
We propose two strategies for HMC of scientific documents.
The first strategy relies solely on LLMs to traverse the taxonomy and select relevant labels.
The second strategy, includes three approaches, combines bi-encoder models for initial filtering, followed by LLM-based refinement of the label selection.
The following subsections provide a detailed breakdown of each approach.
\subsection{LLM-Traverse-LLM-Select (TravSelect)}
In the TravSelect approach, an iterative hierarchical classification process is employed. This involves prompting the LLM to traverse the taxonomy layer by layer using a Breadth-First Search strategy:
\begin{enumerate*}[label=(\roman*)]
    \item \textbf{First Step:} The LLM prompted to evaluate top-level taxonomy nodes to identify relevant categories based on the scientific document's content.
    \item \textbf{Iterative Process:} Each selected node in a layer can either be a leaf, i.e., a node without children, or a parent node. All selected leaf nodes are added to the set of selected nodes. For the selected parent nodes, the LLM continues narrowing down and progressively assessing their children.
    \item \textbf{Final Selection:} The process continues until there is no more parent node among selected nodes, resulting in a set of selected leaf nodes. The prompt
template can be seen in Figure \ref{fig:travselect} in the appendix.
\end{enumerate*}
\subsection{Initial Filtering with Bi-Encoder Models}
This set of approaches begins with a common step: filtering the taxonomy using a bi-encoder model.
This step involves ranking all leaf nodes of the taxonomy based on their similarity to the given scientific document's content.
The bi-encoder model computes the cosine similarity between the embeddings of the scientific document and the taxonomy nodes where each node is represented by its name and description.
The objective is to eliminate irrelevant leaf nodes early, reducing the computational load for subsequent steps.
\par
\begin{figure}[t]
\centering
\includegraphics[width=0.43\textwidth]{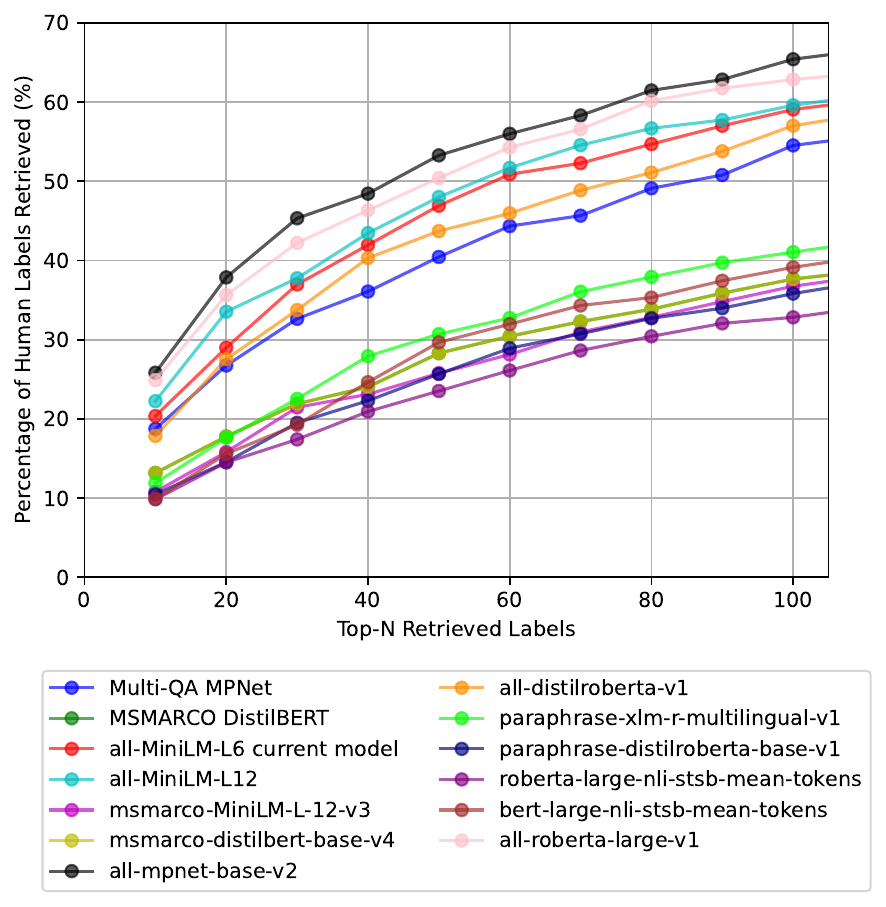}
\caption{Comparative performance of different bi-encoders.}
\label{fig:bi-encoders1}
\end{figure}

In our experiments, we evaluated several bi-encoder models to assess their effectiveness in ranking human-selected labels among the top positions, as shown in Figure \ref{fig:bi-encoders1}. In this setup, we only had only one perfect set of labels for each scientific document.
The "sentence-transformers/all-mpnet-base-v2" model consistently outperformed other models and was thus selected for the initial filtering step in all subsequent approaches. 
We also explored different top-k depths, ranging from 10 to 100.
Consequently, to optimize both effectiveness and computational costs, we present the top 40 leaf nodes, as suggested by SMEs after analyzing the best performing methods results, from the bi-encoder model, along with their hierarchical context (i.e., the full path to the root) as the pruned taxonomy (PT), to our proposed LLM-based classification methods, where each method uses this information differently to select the most relevant labels from this pruned set, considering both the document content and hierarchical relationships.
\subsection{LLM-Based classification methods} \label{sec:llm_classification}
After filtering, each approach differentiates in how it utilizes LLMs for multi-label classification:
\par
\subsubsection{LLM-Select-One-Pass (LLM-SelectO)}

LLM-SelectO adopts a one-pass selection approach, where the LLM is tasked with simultaneously classifying all potential labels in a single prompt, as opposed to individual pointwise classification. The LLM is prompted with the PT, including the description of each label, and tasked with selecting the most relevant labels, considering both the scientific document's content and the hierarchical relationships within the pruned taxonomy. The prompt is presented in Figure \ref{fig:llm_selecto} in appendix.
\par
\subsubsection{LLM-Rerank}
In LLM-Rerank, the LLM is used to assign relevancy scores to the each node from the PT. 
The process involves:
\begin{enumerate*}[label=(\roman*)]
    \item \textbf{Relevancy Scoring:} The LLM assigns a score to each node and its direct parent in the PT based on its similarity to the scientific document used to sort nodes. The prompt is presented in Figure \ref{fig:llm_rerank} in appendix.
    \item \textbf{Re-Ranking:} The scores are then used to rank the taxonomy leaf nodes by applying mathematical functions that consider both the children node scores and their parent nodes.
\end{enumerate*}
The used mathematical functions are as follow: (1) Using only the leaf node’s score; (2) Averaging the score of the leaf node with its direct parent; (3) Averaging the score of the leaf node with all its ancestor nodes; and (4) Using the harmonic mean of the leaf node’s score and those of all its ancestor nodes.
We empirically found that the most effective mathematical function for calculating the final relevance score is the assigned scores to the leaf nodes themselves without considering the parents.
\par
\subsubsection{LLM-Select-Pointwise (LLM-SelectP)}

LLM-SelectP follows a pointwise classification approach, breaking down the HMC task into a series of independent binary classification decisions as illustrated in Figure \ref{fig:selectp}. The process is divided into the following steps:
\begin{enumerate*}[label=(\roman*)]
    \item \textbf{Leaf Node Assessment:} The LLM determines whether each leaf node is relevant based on its description (its prompt is presented in Figure \ref{fig:selectp_label_eval} in appendix);
    \item \textbf{Parent Node Assessment:} The LLM assesses parent nodes to ensure contextual relevance within the hierarchy (its prompt is presented in Figure \ref{fig:selectp_parent_eval} in appendix);
    \item \textbf{Label Adjustment:} The number of selected labels is adjusted to meet the predefined range, ensuring sufficient but not excessive label assignment.
\end{enumerate*}
\par
\subsection{Post-Processing}
All approaches conclude with a post-processing step to refine the final label set, a recommendation from SMEs. This step is highly task-dependent and tailored to the specific requirements of the given problem.
\begin{enumerate*}[label=(\roman*)]
    \item \textbf{Decreasing the Number of Labels:} If more than five labels are selected, the label set will be reduced. The LLM is provided with the selected nodes and their parents and is prompted to choose the most relevant five labels, ensuring the number of labels per document remains within the preferred range, the prompt presented in Figure \ref{fig:postprocess_decreasing} in appendix. This is not applied for LLM-Rerank method where the labels are already scored and top-k labels could be selected straightforwardly.
    \item \textbf{Decreasing number of siblings.} This step is based on SME's suggestion and the goal is to ensure that not all labels are selected from one parent; preventing from being biased to a single subcategory within the taxonomy.
\end{enumerate*}
\begin{table*}[]
\small
\centering
\caption{Effectiveness results of different methods. Machine learning based method \cite{singh2022scirepeval} is the previous SOTA on this task. S-i\% refers to the percentage of the documents that are scored to $i$ by SME for a method. SelectL and SelectP refers to Listwise and Pointwise respectively.}
\label{tab:results}
    \scalebox{0.995}{
        \begin{tabular}{p{5cm}|c|c|c|c|c|c} 
            \toprule
            Method                             & Accuracy\% & S-5\% ↑ & S-4\% ↑ & S-3\% ↑ & S-2\% ↓ & S-1\% ↓  \\ \midrule
            \multicolumn{7}{c}{Machine learning based method} \\
            \midrule
            Previous SOTA   \cite{singh2022scirepeval}                     &    61.5      &  00.0  &  11.5  &  50.0  &  30.7  &  7.8  \\
            \midrule
            \multicolumn{7}{c}{Only LLM-based method} \\
            \midrule
            Trav-Select   (ours)                      &    50.0      &  4.3  &  14.3  &  25.7  &  22.9  &  32.9  \\
            \midrule
            \multicolumn{7}{c}{Bi-encoder followed by LLM-based methods} \\
            \midrule
            LLM-Rerank  (ours)                      &    70.0      &  0.0  &  4.3  &  \textbf{60}  &  31.4  &  4.3  \\
            LLM-SelectO (ours)                    &    58.6      &  4.3  &  24.3  &  25.7  &  28.6  &  17.1  \\
            LLM-SelectP (ours)              &    \textbf{94.3}      &  \textbf{32.9}  &  \textbf{38.6}  &  22.9  &  \textbf{4.3}  &  \textbf{1.4}  \\
            \midrule
            \multicolumn{7}{c}{Ablation analysis(Ours)} \\
            \midrule
            LLM-SelectP w/o decreasing (random selection) & 62.9\ & 0.0\ & 4.3\ & 50.0\ & 37.1\ & 8.6\ \\
            LLM-SelectP w/o description & 85.7\ & 2.9\ & 15.7\ & 60.0\ & 18.6\ & 2.9\ \\
            LLM-SelectP w/o contextualization & 85.7\ & 2.9\ & 28.6\ & 57.1\ & 7.1\ & 4.3\ \\
            \bottomrule
        \end{tabular}
    }
\end{table*}
\section{Evaluation Framework}
\label{Evaluation Framework}
Given the possibility of having multiple perfect label sets for each document, we could not rely on a gold dataset for evaluation. Instead, we engaged SMEs to provide direct feedback on the labels assigned by each method to scientific documents. SMEs reviewed a set of 100 documents for each method, evaluating the accuracy and relevance of the assigned labels.
To evaluate the HMC models, we used two metrics:
\begin{enumerate*}[label=(\roman*)]
    \item \textbf{Correctness}: A binary metric indicating whether the SME deemed the selected label set by the appropriate. We report the percentage of positive responses as accuracy.
    \item \textbf{Subjective Evaluation}: SMEs rated label quality on a 1-5 scale. The detailed explanation of scores is given in Table \ref{tab:scores-explanation} in Appendix.
\end{enumerate*}
We used the GENEX, an evaluation tool developed by Elsevier (Figure \ref{fig:genex_schema}), to assist SMEs in evaluating the labels and gathering quantitative feedback, including questions like "What is the ideal label set?", "Why did you assign this score?", and "What makes a label set unsuitable?" These insights were pivotal during the Proof of Concept (PoC) phase to address approach limitations.
\section{Results}
\label{sec:results}
Table \ref{tab:results} presents accuracy and SME scoring metrics (S-1\% to S-5\%), which represent the percentage of documents rated from 1 (unacceptable quality) to 5 (excellent quality). The results show that our proposed methods, which combine a bi-encoder for initial filtering and classification by LLMs, outperform the previous SOTA, SPECTER2 \cite{singh2022scirepeval}. Our best method, LLM-SelectP, achieves an accuracy of 0.943 compared to 0.615 for SPECTER2. Furthermore, LLM-SelectP, by a large margin, achieves the highest effectiveness in terms of S-5\%, with 32.9\% of its predictions rated as perfect annotations, while the previous SOTA achieves 0\% in this setup. Even other proposed methods are limited to 4.3\% of predictions rated as perfect annotations.
We also found that having the LLM approach the classification task alone, as in the Trav-Select method, results in lower effectiveness compared to all proposed methods and the previous SOTA. These results underscore the importance of effective initial label selection, particularly for large taxonomies.
\par
\header{Ablation Analysis}
We analyzed the importance of each component of LLM-SelectP's full methodology. Table \ref{tab:results} shows skipping decreasing the number of labels reduced performance significantly with a drop of 32\% in terms of accuracy. Furthermore, removing label descriptions where we only provide label title without its description and without contextualization where we skip evaluation of parent node results in a drop of about 9\% in terms of accuracy indicating the importance of all of these steps in LLM-SelectP method.
\par
\section{Business Impact}
The proposed AI Classification system implemented for SSRN, Elsevier’s preprint repository, has fundamentally transformed the process of document categorization. Prior to this, human classifiers manually assigned over 3,000 constantly evolving labels, which became increasingly impractical due to growing business demands and the rapid expansion of academic disciplines.
By automating this process using ChatGPT 3.5, SSRN now classifies documents in a fraction of the time and at a fraction of the cost. Each manually classified document previously cost approximately \$3.50, while our system processes them for around \$0.20 each.
With over 140,000 papers submitted annually, this reduction in classification costs results in substantial financial savings, projected to exceed \$100,000 in 2024 alone. This transformation allows SSRN to redirect resources towards strategic initiatives, ensuring scalability and sustained operational efficiency as the volume of submissions grows.
The system runs daily, eliminating the backlog that once delayed the processing of papers, and providing a consistent quality that surpasses the accuracy of manual classification. As SSRN scales, this AI-driven approach ensures that both cost and operational bottlenecks are mitigated, freeing up resources for more strategic initiatives and allowing SSRN to keep pace with the rapidly evolving academic landscape.
\section{Conclusion}
\label{Conclusion}

In this paper, we present novel approaches for HMC of scientific documents using LLMs and dense retrievers. Our methods, without the need for training, effectively handle large, dynamic taxonomies. Among the various approaches we proposed, the LLM-SelectP method achieved over 94\% effectiveness in terms of accuracy, highlighting the potential of LLMs in large-scale, real-world classification tasks.

While our current approach successfully utilizes document metadata (title, abstract, and keywords) to maintain cost-effectiveness, future work could explore the integration of full-text analysis, particularly for cases where the system shows lower confidence in classification. Our decision to exclude full-text processing was primarily driven by cost considerations, as LLM processing costs typically scale with token count. However, a hybrid approach that selectively processes full text for ambiguous cases could potentially further improve accuracy while maintaining reasonable operational costs.

\bibliography{coling_latex.bib}
\clearpage
\appendix
\section{Appendix}
\subsection{Explanation of Quality Scores for Classification}
In this section, we explain the quality scores used in evaluating the classifications. Each score corresponds to a specific level of classification quality, ranging from unacceptable to excellent. The score descriptions focus on the presence of essential classifications, the occurrence of wrong or low-value classifications, and the overall impact on the discovery experience for researchers and the satisfaction of authors. The scores are defined at Table \ref{tab:scores-explanation}.
\begin{table*}[]
\small
\centering
\caption{Quality scores for classification.}
\label{tab:scores-explanation}
\begin{tabular}{l|p{15cm}} 
\toprule
\textbf{Score} & \textbf{Explanation} \\ 
\midrule
1 & \textbf{Unacceptable Quality.} All essential classifications are missing. \\ 
\midrule
2 & \textbf{Low Quality. }One or more essential classifications are missing, minimal relevant classifications are present, and more than 1 classification is wrong. The low quality would prevent a good discovery experience for researchers and would irritate authors. \\ 
\midrule
3 & \textbf{Acceptable Quality.} No essential classifications are missing, at most one classification is wrong, and there may be some relevant classifications. Overall, this quality enables a decent discovery experience, satisfactory for most authors, and is nearly as good as a human classifier would provide. \\ 
\midrule
4 & \textbf{Good Quality.} All essential classifications are present, no classifications are wrong, and minimal low-value classifications exist. The quality supports discovery across disciplines and matches what we would expect from a human classifier, providing a good discovery experience that most authors would welcome. \\ 
\midrule
5 & \textbf{Excellent Quality.} All essential classifications are present, with no low-value or wrong classifications. Overall, the classifications match or exceed the quality of human classifiers, offering an excellent discovery experience that will please researchers and impress authors. \\
\bottomrule
\end{tabular}
\end{table*}

\subsection{Subjectivity of Annotation}
\begin{figure*}[h!]
    \centering
    \scalebox{0.98}{\includegraphics[width=1\linewidth]{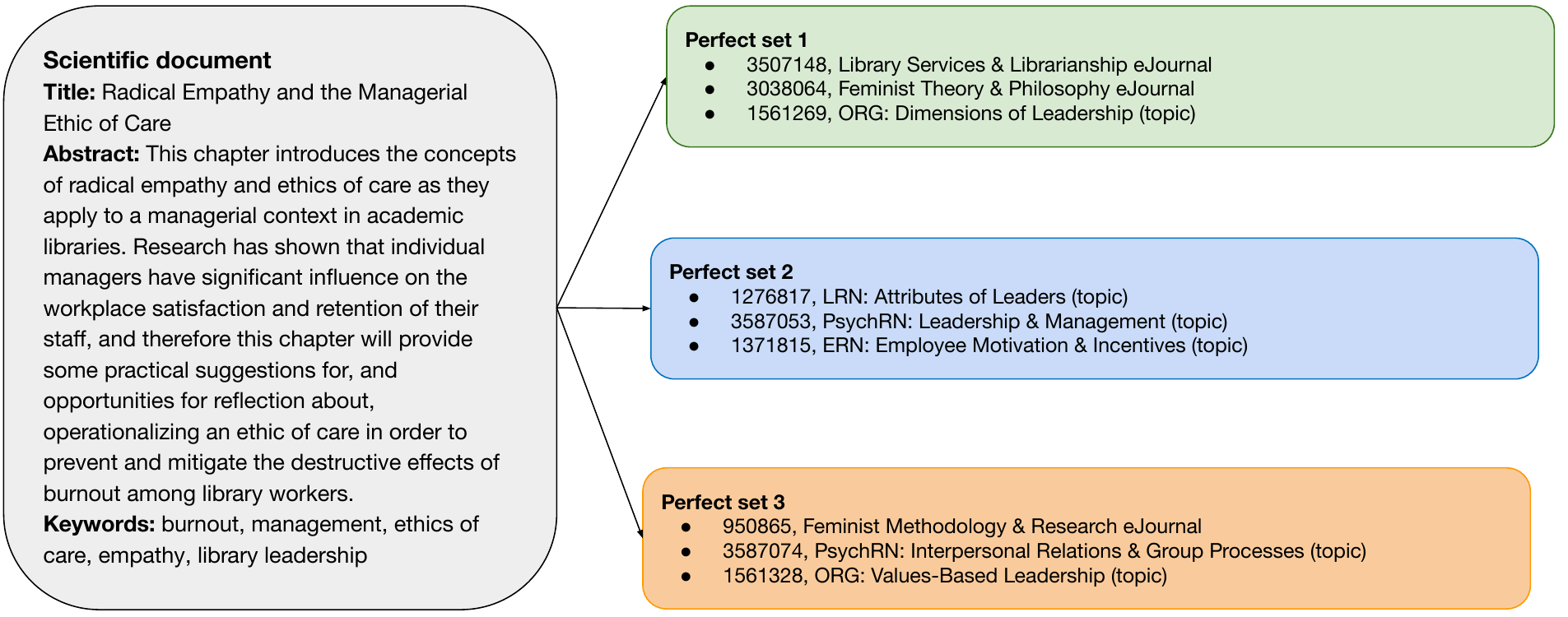}}
    \caption{A document can belong to multiple perfect sets, each consisting of different combinations of relevant labels.}
    \label{fig:perfect_sets}
\end{figure*}

Figure \ref{fig:perfect_sets} illustrates an example of a scientific document with three different sets of labels, each of which could be considered a perfect match for the document. This highlights the inherent subjectivity of the task, as multiple label sets can be deemed ideal for the same document. Consequently, this necessitates a dynamic evaluation approach tailored to each method.

\subsection{Previous Solution: Human Classifiers}
This section outlines some key challenges with the human classification system and the limitations of the current taxonomy structure, which has impacted the quality and consistency of classification over time.
\subsubsection{Human Classifier Limitations}
Several factors contribute to the varying levels of quality in the classification performed by human classifiers:
\begin{itemize}
    \item \textbf{Part-time nature of the role}: SSRN classifiers are typically part-time contract workers, many of whom have other professional obligations. Until recently, the hourly wage was quite modest (only \$15 per hour), meaning that for some, the position was not a high-priority role. Consequently, there has been limited motivation to perform the job exceptionally well.
    \item \textbf{Lack of incentives}: Compensation for the classification work has not been directly tied to either speed or quality. Historically, there were no financial incentives such as pay raises for consistently high-quality work. This has led to varying levels of engagement and output quality across classifiers.
    \item \textbf{Cumbersome workflow}: The classification process has been organized around "networks," each with separate queues and individual classifiers. Due to this structure, a paper may be examined by different people, each responsible for classifying within a specific section of the taxonomy. This fragmented approach leads to inconsistencies in classification across different networks, as there is no unified process for adding all relevant classifications at once. Additionally, errors made by front-end processors (often low-wage workers without advanced subject matter knowledge) can result in papers being omitted from relevant queues, further compounding the inconsistency.
\end{itemize}
\subsubsection{Taxonomy Structure Challenges}
The structure and evolution of the taxonomy itself has also contributed to classification challenges:
\begin{itemize}
    \item \textbf{Siloed taxonomies}: The current taxonomy system has developed over approximately 30 years, and was historically built in isolation across different networks. This has resulted in overlapping yet functionally separate silos (e.g., Cognitive Science, Neuroscience, and Decision Science) that conceptually overlap but are treated as distinct workflows. Only recently has there been an effort to integrate these taxonomies into a unified system and develop a holistic view of classification.
    \item \textbf{Inconsistent terminology and duplication}: Due to the historical isolation of taxonomies, there have been issues with overlap, inconsistent terminology, and duplication across categories. Furthermore, not all subject areas have a suitable label in the current taxonomy, which can lead to errors of omission during classification.
    \item \textbf{User-driven taxonomy}: The existing taxonomy has also been shaped by user demand, particularly through subscriber-driven email alerts. As a result, some labels are extremely broad (e.g., "Ecology eJournal"), while others are more niche (e.g., "Law, Policy, \& Economics of Technical Standards eJournal"). This demand-driven approach has not always aligned with the subject matter itself, complicating the classification process.
\end{itemize}

\subsection{Prompts}
\subsection{Description Generation}
The prompt of our description generation is presented in Figure \ref{fig:descgen}.

\begin{figure}[h]
    \begin{mdframed}[backgroundcolor=verylightgray,roundcorner=10pt]
    \begin{lstlisting}
You are an AI assistant designed to generate descriptions for labels used in classifying SSRN preprint articles. To do this, you should use the information in the name of the label, and also using the information about the parent of the label in the taxonomy.
    \end{lstlisting}
    \end{mdframed}
    \caption{The prompt of description generation.}
    \label{fig:descgen}
\end{figure}

\subsection{Traverse Prompt}
The prompt of our LLM-Traverse-LLM-Select (TravSelect) method is presented in Figure \ref{fig:travselect}.
\begin{figure}[h]
    \begin{mdframed}[backgroundcolor=verylightgray,roundcorner=10pt]
    \begin{lstlisting}
You are an AI trained to evaluate the relevance of multiple labels for a given SSRN pre-print document. For this task, you will receive the document's title, keywords, abstract, and a list of labels. Each label in the list has an ID, a name, and description. Your task is to determine which labels are the best fit for the document. A label fits well if the document's main focus aligns with the area the label describes. Your output should be a concise JSON object containing a list, 'best_labels', which only includes the ID of labels that best fit the document.
    \end{lstlisting}
    \end{mdframed}
    \caption{The prompt of LLM-Traverse-LLM-Select (TravSelect) method.}
    \label{fig:travselect}
\end{figure}

\subsection{LLM-Select-One-Pass Prompt}
The prompt of LLM-Select-One-Pass (LLM-SelectO) method is presented in Figure \ref{fig:llm_selecto}.
\begin{figure}[h]
    \begin{mdframed}[backgroundcolor=verylightgray,roundcorner=10pt]
    \begin{lstlisting}
You are an AI assistant trained to evaluate the relevance of multiple labels for a given SSRN pre-print document. You will receive the document's title, keywords, abstract, and a taxonomy of labels. Each label in the taxonomy has an ID, a name, and description. Your task is to select the best-fitting leaf labels (having no children) for the document.
A label is considered a good fit if:
- It directly relates to the core subject of the article.
- All its parents are relevant to the document.
Your output should be a concise JSON object containing a list, 'best_labels', which only includes the IDs of the labels that best fit the document.
    \end{lstlisting}
    \end{mdframed}
    \caption{The prompt of LLM-Select-One-Pass method.}
    \label{fig:llm_selecto}
\end{figure}

\subsection{LLM-Rerank Prompt}
The prompt of our LLM-Rerank method is presented in Figure \ref{fig:llm_rerank}.

\begin{figure}[h]
    \begin{mdframed}[backgroundcolor=verylightgray,roundcorner=10pt]
    \begin{lstlisting}
You are an AI assistant helping me to find the conceptual similarity scores between an SSRN article and a list of {} labels.  
Please ensure the following:
- Return a score for each label.
- Ensure there are {} scores in total.
- Ensure the scores are varied and accurately represent the level of similarity, rather than scoring a large percentage of labels the same.
- Consider the main theme of the article and the specific context in which keywords are used.
- Do not assign high similarity scores to labels that are only tangentially related or share a few keywords with the article. The focus should be on the overall subject matter of the article.
- Scores should have two decimal points for greater precision.
The output should be a JSON object named "scores" that contains a list of {} tuples. Each tuple should contain a label ID and a relevancy score between 0.01 and 1.00, indicating the level of relevancy between the label and the given document.
    \end{lstlisting}
    \end{mdframed}
    \caption{The prompt of LLM-Rerank method.}
    \label{fig:llm_rerank}
\end{figure}

\subsection{LLM-Select-Pointwise Prompts}
The prompts for the leaf label and parent label assessments in the LLM-Select-Pointwise method are shown in Figures \ref{fig:selectp_label_eval} and \ref{fig:selectp_parent_eval}.
\begin{figure}[h]
    \begin{mdframed}[backgroundcolor=verylightgray,roundcorner=10pt]
    \begin{lstlisting}
You are an AI trained to evaluate the relevance of a label for a given SSRN pre-print document. You will receive the document's title, keywords, abstract, and the label's ID, name, and description. Your task is to determine if the label is a good fit for the document. A label fits well if the document's main focus aligns with the area the label describes. Your output should be a concise JSON object. The JSON object should contain three keys: "main_focus", a very short representation of the document's main focus, "label_fit", representing the fit as a boolean value. It's crucial to utilize the entire scoring range to reflect varying degrees of relevancy. Please do not provide any further information or explanation in addition to the JSON object. Do not use the slash or backslash characters in your output.
    \end{lstlisting}
    \end{mdframed}
    \caption{The prompt of LLM-Select-Pointwise method for the leaf label assessment.}
    \label{fig:selectp_label_eval}
\end{figure}
\begin{figure}[h]
    \begin{mdframed}[backgroundcolor=verylightgray,roundcorner=10pt]
    \begin{lstlisting}
You are an AI, trained to assess the potential relevance of a label for a given SSRN pre-print document. You'll be provided with the document's title, keywords, abstract, and the label's name and description. Your mission is to gauge if the label could be a reasonable match for the document. A label can be considered a reasonable match even if it only partially aligns with the document's main theme. Your response should be a JSON object. This JSON object should include three keys: "main_focus", a brief summary of the document's main theme, "label_fit", indicating the fit as a boolean value, and "relevancy_score", showing the relevance as a score from 0 to 1. It's important to use the full scoring range to indicate varying levels of relevance. Do not use the slash or backslash characters in your output.
    \end{lstlisting}
    \end{mdframed}
    \caption{The prompt of LLM-Select-Pointwise method for the parent label assessment.}
    \label{fig:selectp_parent_eval}
\end{figure}
\subsection{Decreasing the number of labels}
The prompt for decreasing the number of labels in post-processing is presented in Figure \ref{fig:postprocess_decreasing}.
\begin{figure}[h]
    \begin{mdframed}[backgroundcolor=verylightgray,roundcorner=10pt]
    \begin{lstlisting}
You are an AI trained to evaluate the relevance of multiple labels for a given SSRN pre-print document and select the top 5 labels that best fit the document. For this task, you will receive the document's title, keywords, abstract, and a list of labels. Each label in the list has an ID, name, and description. Your task is to determine which labels are the best fit for the document. A label fits well if the document's main focus aligns with the area the label describes. Please return the IDs of the top 5 labels that best fit the given document.
    \end{lstlisting}
    \end{mdframed}
    \caption{The prompt for decreasing the number of labels in post-processing.}
    \label{fig:postprocess_decreasing}
\end{figure}
\begin{figure*}[h]
    \centering
    \scalebox{1}{\includegraphics[width=1\textwidth]{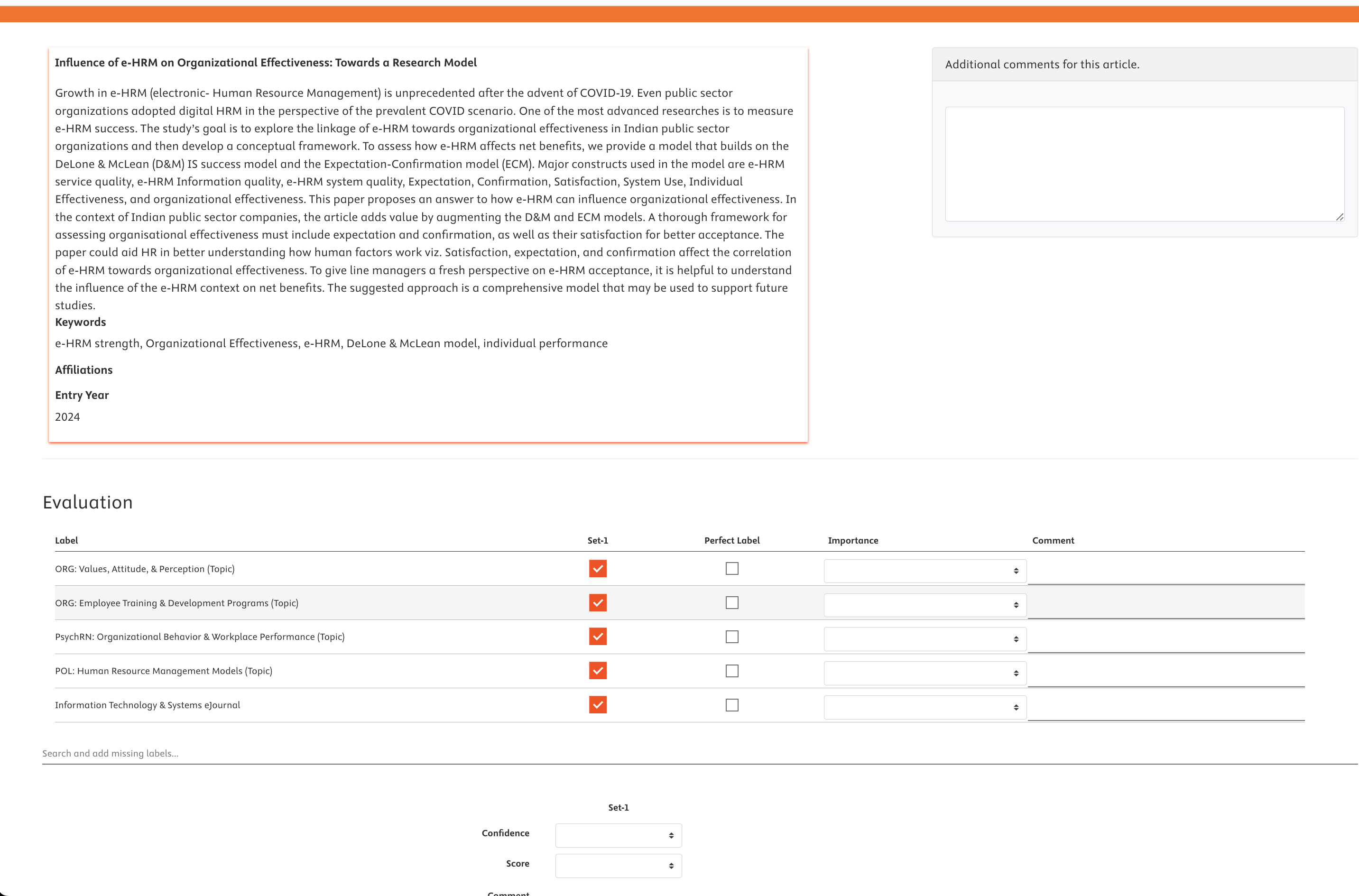}}
    \caption{Schema of the GENEX tool used for evaluation.}
    \label{fig:genex_schema}
\end{figure*}

\end{document}